\documentclass{article}
\usepackage{ijcai23}

\usepackage{times}
\usepackage{soul}
\usepackage{url}
\usepackage[hidelinks]{hyperref}
\usepackage[utf8]{inputenc}
\usepackage[small]{caption}
\usepackage{graphicx}
\usepackage{amsmath}
\usepackage{amsthm}
\usepackage{booktabs}
\usepackage{algorithm}
\usepackage{algorithmic}
\usepackage[switch]{lineno}

\usepackage[dvipsnames]{xcolor}
\newif\ifmodify
\ifmodify

\newcommand{\xs}[1]{\textcolor{blue}{#1}}
\newcommand{\cross}[1]{\textcolor{gray}{\sout{#1}}}
\newcommand{\todo}[1]{\textcolor{red}{#1}}
\else
\newcommand{\cross}[1]{}
\newcommand{\todo}[1]{}

\newcommand{\xs}[1]{#1}

\fi

\usepackage{makecell}
\usepackage{multirow}
\usepackage{multicol}
\usepackage{booktabs}
\usepackage{enumitem}
\usepackage{pifont}
\usepackage[normalem]{ulem}
\usepackage{amssymb}

\title{Data Level Lottery Ticket Hypothesis for Vision Transformers}

\author{
Xuan Shen$^1$
\and
Zhenglun Kong$^1$\and
Minghai Qin$^{2}$\and
Peiyan Dong$^{1}$\and
Geng Yuan$^{1}$\and
Xin Meng$^{3}$\and \\
Hao Tang$^{4}$\and
Xiaolong Ma$^{5}$\And
Yanzhi Wang$^1$
\affiliations
$^1$Northeastern University \\
$^2$Western Digital Research \\
$^3$Peking University \\
$^4$ETH Zurich \\
$^5$Clemson University \\
\emails
\{shen.xu, kong.zhe, dong.pe, yuan.geng, yanz.wang\}@northeastern.edu, 
qinminghai@gmail.com,
1601214372@pku.edu.cn,
hao.tang@vision.ee.ethz.ch,
xiaolom@clemson.edu,
}
\begin{document}

\maketitle

\begin{abstract}

The conventional lottery ticket hypothesis (LTH) claims that there exists a sparse subnetwork within a dense neural network and a proper random initialization method called the {\em winning ticket}, such that it can be trained from scratch to almost as good as the dense counterpart.
Meanwhile, the research of LTH in vision transformers (ViTs) is scarcely evaluated.
In this paper, we first show that the conventional winning ticket is hard to find at the weight level of ViTs by existing methods.
\xs{Then, we generalize the LTH for ViTs to input data consisting of image patches inspired by the input dependence of ViTs.}
That is, there exists a subset of input image patches such that a ViT can be trained from scratch by using only this subset of patches and achieve similar accuracy to the ViTs trained by using all image patches. We call this subset of input patches the {\em winning tickets}, which represent a significant amount of information in the input data.
We use a ticket selector to generate the winning tickets based on the informativeness of patches for various types of ViT, including DeiT, LV-ViT, and Swin Transformers.
The experiments show that there is a clear difference between the performance of models trained with winning tickets and randomly selected subsets, which verifies our proposed theory.
We elaborate on the analogical similarity between our proposed Data-LTH-ViTs and the conventional LTH to further verify the integrity of our theory.
The Source codes are available at \textcolor{magenta}{\url{https://github.com/shawnricecake/vit-lottery-ticket-input}}.

\end{abstract}

\section{Introduction}
In recent years, computer vision research has evolved into Transformer architectures since Vision Transformer (ViT)~\cite{dosovitskiy2020vit} achieved excellent performance on image classification tasks. Models such as DeiT~\cite{pmlr-v139-touvron21a-deit} and Swin Transformer~\cite{liu2021Swin} have achieved state-of-the-art performance of image classification on CIFAR-10/100~\cite{krizhevsky2009learning} and ImageNet~\cite{deng2009imagenet} datasets. As great attention has been drawn to the ViT, computation redundancy becomes an issue because of the sacrifice on the lightweight model capacity for higher accuracy. Based on~\cite{dosovitskiy2020vit}, training a ViT from scratch requires large datasets (i.e., ImageNet), and more iterations are required for the convergence of the ViT to ensure that the network is fully exploited. The large dataset and complex model architecture motivate us to explore the lottery ticket hypothesis for efficient training and testing (inference) of ViT models. 

In parallel, the Lottery Ticket Hypothesis (LTH)~\cite{frankle2018lottery} suggests that there exists some {\em winning tickets} (i.e., a properly pruned subnetwork together with the original weight initialization) that can achieve performance competitive with that of the original dense network. Such winning tickets in CNNs can achieve comparable performance to the dense network on image classification tasks and allow the training of the network to be more efficient. Since then, LTH has been applied to various research areas like BERT and object detection~\cite{chen2020lottery,prasanna2020bert,bai2022dual-lottery,desai2019evaluating,yu2019playing}. 
Nonetheless, the existence of winning tickets of ViTs is not explored by the conventional LTH definition.

Recent works on LTH  have focused on identifying the winning tickets at the weight level in DNNs. 
We follow the settings of conventional LTH~\cite{frankle2018lottery} in the experiments on ViTs.
Interestingly, the experimental results at the weight level on DeiT-Small~\cite{pmlr-v139-touvron21a-deit} indicate that
it is hard to find such winning tickets that make training effective even with the help of the rewinding technique.
Details of these experiments can be found in Table~\ref{weight-lth-1} and Table~\ref{weight-lth-2}. The absence of the winning tickets is
mainly due to the input dependence of the ViT, as it is difficult to identify a subnetwork at the weight level to generalize to the different input images.
\xs{The difficulty of finding weight-level winning tickets drives us to find them on the data level.}

In ViTs, each image is first split into square patches, which are then projected to tokens before being fed into the transformer blocks.
Recent works~\cite{rao2021dynamicvit,xu2022evovit,liang2022evit,kong2022spvit} demonstrate a series of dynamic token pruning strategies, where the tokens are pruned independently in the subnetworks based on the features of each input image. 
These observations motivate us to explore the LTH at the data level.
\xs{We are supposed to identify the winning tickets on the input data instead of model weights or tokens. As the input data for ViTs is the image patches and many works~\cite{liang2022evit,kong2022peeling,rao2021dynamicvit} show that the different image patches have different attentiveness and informativeness, we can build the strategy of identifying winning tickets according to those attributes of input data.}

\begin{figure}[t]
    \centering
    \includegraphics[width=0.47\textwidth]{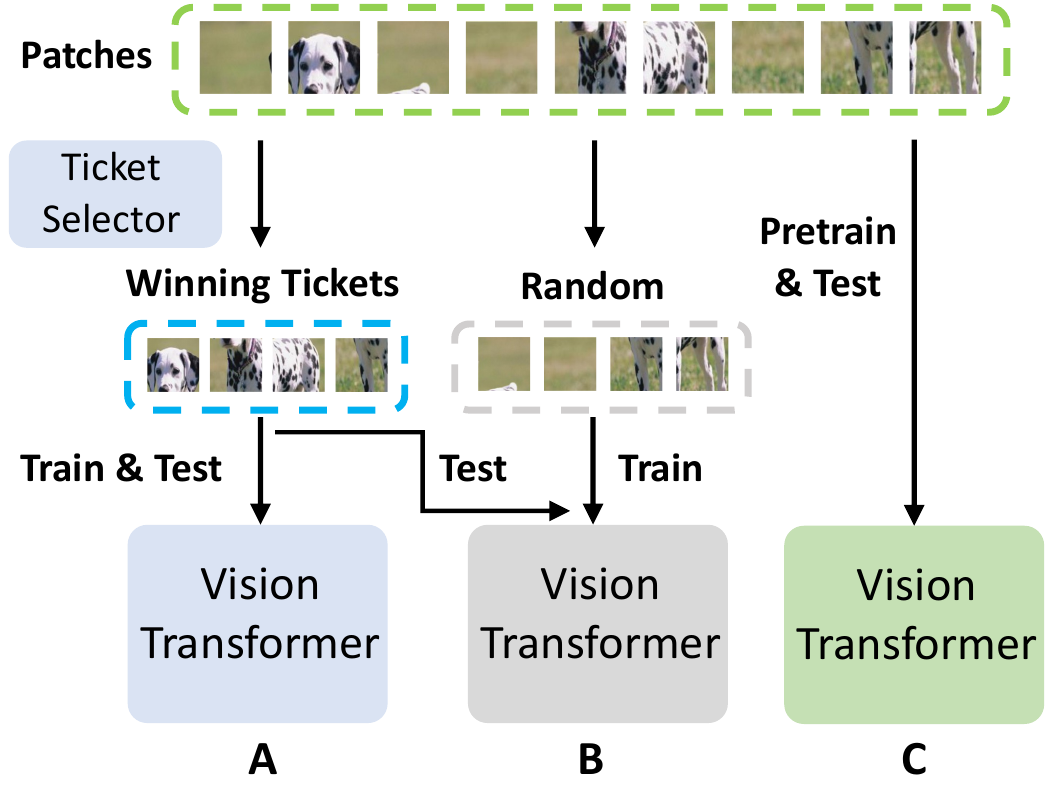}
    \caption{Overview of the \xs{Data Level} Lottery Ticket Hypothesis for Vision Transformers. Three paths are introduced for a complete definition: (a) the Lottery Tickets are generated by the ticket selector, and they are subsets of input image patches that a ViT is trained from scratch and tested, (b) random input patches are selected by the random selector and are used to train a ViT, while the same ticket selector is used to generate a subset of patches for the test, (c) all image patches are used to train and test a ViT. Our \xs{data level} LTH claims that (a) $\approx$ (c) $>$ (b) in accuracy, with the notable margin between (a) and (b).}
    \label{figure-overview-of-training-testing-pipeline}
\end{figure}

In this paper, we propose a new \xs{Data Level} Lottery Ticket Hypothesis for Vision Transformers (\xs{Data-}LTH-ViTs), shown in Figure~\ref{figure-overview-of-training-testing-pipeline}.
In the first path, (a), for each input image, we use a ticket selector to produce the subset of attentive image patches before feeding them into the ViT for both training and testing.
The ticket selector removes input patches with less contribution according to the attention map in transformer blocks. The positional embeddings of the remaining tokens further guide the identification of attentive image patches for training.
Then, the model is trained and tested with the subset of selected patches, identified as {\em winning tickets}.
The second path, (b), selects random input image patches for training. In order to compare  with Path (a), we use the same ticket selector to generate input patches for testing.
The last path, (c), is a normal ViT training and test process where all input patches are used for training and testing.
The experiments we conducted in this paper show that our method of training the ViTs only with winning tickets (Path (a)) can match the performance of conventionally trained ViT models (Path (c)) and have notable accuracy advantages over ViTs trained by a random subset of input patches (Path (b)). Based on this observation, we propose the LTH for ViTs, and it suggests that there exist winning tickets at the input image patch level that can make the training and test more efficient and effective.
To compensate for our proposed \xs{data level} LTH for ViTs, we also try to identify the winning tickets on CNNs with our proposed method.
Interestingly, it is difficult for input-independent CNNs to identify the winning tickets with the proposed method. The results can be found in Table~\ref{table-other-kinds-of-ViT}.

% \textbf{Analogical similarity between conventional LTH and the proposed \xs{Data-}LTH-ViTs.}
% Both conventional LTH and \xs{Data-}LTH-ViTs 
% \ding{172} attempt to identify the subsets of weights/inputs that can achieve similar accuracy to the model trained with full weights/inputs sets;
% \ding{173} need to use prior knowledge (i.e. a pretrained model) to identify the winning tickets;
% \ding{174} identify the winning tickets with the requirement of matching the accuracy of pretrained model;
% \ding{175} import randomness into the subsets to validate the uniqueness and effectiveness of the winning tickets.

\xs{We first summarize the analogical similarity between conventional LTH and the proposed Data-LTH-ViTs as follow:}

\begin{itemize}%[leftmargin=*, noitemsep,topsep=0pt]

\item 
Both theories attempt to identify the subsets of weights/data that can achieve similar accuracy to the model trained with full weights/data sets.

\item
Both theories need to use prior knowledge (i.e., a pretrained model) to identify the winning tickets.

\item
Both theories identify the winning tickets with the requirement of matching the accuracy of pretrained model.

\item
Both theories import randomness into the subsets to validate the uniqueness and effectiveness of the winning tickets.

\end{itemize}

We \xs{then} summarize our contributions as follows:

\begin{itemize}%[leftmargin=*, noitemsep,topsep=0pt]

\item 

We show by experiments that the conventional winning ticket is hard to find \xs{in ViTs} even with the rewinding technique.
In particular, there is a negligible difference in accuracy between a random sparse weight initialization and the proper initialization defined in the conventional LTH.

\item
We generalize the definition of winning tickets in LTH from sparse weights to \xs{sparse data}. That is, there exists a subset of \xs{input data} such that a ViT can be trained from scratch by using only this subset of data and achieve similar accuracy to the ViTs trained by full data. This \xs{subset of input data} in training is called the {\em input winning ticket} of ViTs.

\item
We propose a method to identify the winning tickets at the data level for ViTs.
Experiments show that we can reduce 48.8\% of input \xs{data} during training,
which can match the accuracy to the ViTs trained by using \xs{full data}.
We also observe another feature of \xs{data level} winning tickets, which is that their universality across different types of ViTs.

\end{itemize}

\section{Related Works}

\subsection{The Lottery Ticket Hypothesis}
The lottery ticket hypothesis is first proposed by the work~\cite{frankle2018lottery}. The hypothesis suggests that winning tickets found in CNN on CIFAR-10/100~\cite{krizhevsky2009learning} dataset have connections with the initial weights that can make the training particularly effective. The following work~\cite{renda2020comparing-rewinding} extends (rewinds) the training of the subnetwork from initialization to the early stage of pretraining, which improves the accuracy of the subnetwork in more challenging tasks. Furthermore, the work~\cite{frankle2020linear} demonstrates the stability of those subnetworks to the SGD noise in the early stage of training, which explains the success of the rewinding technique in LTH. Besides, the current work~\cite{ma2021sanity} claims that the winning tickets are difficult to find on ImageNet~\cite{deng2009imagenet} dataset for CNNs, and this work also shows the effect of the enough-trained original network on the identically initialized subnetwork.

Apart from the image classification tasks, the LTH is also imported in many other research areas~\cite{chen2020lottery,mallya2018piggyback,gale2019state,yu2019playing,renda2020comparing-rewinding,chen2020lottery,prasanna2020bert,girish2020lottery}. The works~\cite{gale2019state,yu2019playing,renda2020comparing-rewinding} show that the subnetworks exist early in the training instead of initialization on Transformers. Also, the works~\cite{chen2020lottery,prasanna2020bert} study the LTH in BERT models and find the matching subnetworks at high-level sparsity on the pretrained initialization. Besides, the work~\cite{girish2020lottery} proposes a guidance for the identification of task-specific winning tickets for object detection, instance segmentation, and keypoint estimation. However, none of the existing LTH literature explored sparse winning tickets for ViTs, which have been widely utilized in computer vision tasks in recent years.

\subsection{Redundancy in ViTs}
% \noindent\textbf{Redundancy in ViTs.}
The work~\cite{dosovitskiy2020vit} first investigates the Transformers in computer vision and then builds the Vision Transformer (ViT) by dividing one image into several patches. Since then, many variants have been proposed~\cite{wu2021rethinking,chen2021autoformer,steiner2021train,el2021xcit,liu2021efficient,wang2021kvt,bao2021beit}, and
there are several works exploring the redundancy of ViTs from different directions. \cite{yu2022unified,chen2022principle,hou2021multi,chen2021chasing,yu2022mia,pan2021scalable,liang2022evit,yu2022unified,rao2021dynamicvit} apply structured neuron pruning or unstructured weight pruning. \cite{tang2021patch,pan2021ia,xu2021evo} apply dynamic or static token sparsification.
The results achieved by reducing the number of tokens involved in ViTs would be significant compared to those efforts that attempt to reduce the complexity of the Transformer structure with artificially designed modules. DynamicViT~\cite{rao2021dynamicvit} introduces a method to reduce the number of tokens in a well-trained ViT with an extra learnable neural network that is used to generate the subset of the tokens. EViT~\cite{liang2022evit} provides a token reorganization method that reduces and reorganizes the tokens with no additional parameters and achieves good results.
Therefore, based on these works, the token redundancy in ViTs proves to be large and each token can be connected to the corresponding input image patch based on the positional embedding, which means we can achieve token pruning by removing the corresponding image patch. However, these approaches typically need to progressively reduce the number of tokens because ViTs are considered difficult to remove the token at the very beginning due to the loss of performance.

In our work, we propose an effective framework to remove the inattentive image patches before feeding them into the transformer blocks.
Our work is different from the token pruning works that feed the transformer with a full image and prune the tokens between the transformer blocks progressively.
The results show that removing inattentive image patches from training can still match the accuracy of the pretrained model.
Furthermore, the remaining attentive image patches are proved to be effective according to the experimental results of the randomness counterpart and defined as winning tickets in our proposed special \xs{data level} lottery ticket hypothesis in ViTs.
This is the first time that the LTH achieves positive results in vision transformer architectures.
The proposed winning tickets are different from the conventional LTH, whose winning tickets are subnetworks.

\section{The \xs{Data Level} Lottery Ticket Hypothesis for Vision Transformers} \label{sec:LTH}

Our work is mainly based on the DeiT~\cite{pmlr-v139-touvron21a-deit}. We first show that it is hard to find conventional winning tickets in sparse ViTs even with the rewinding technique, and then present how to identify the winning tickets with the analysis of the Transformer architectures. The rigorous definition of the \xs{data level} lottery ticket hypothesis is elaborated and the specific methods are illustrated to demonstrate the existence and effectiveness of the winning tickets in ViTs. Furthermore, other variants of ViTs (e.g., LV-ViT and Swin Transformer) are imported for the verification of the generalization of the winning tickets, and the respective methods are adjusted to adapt to the LTH experiments.

\subsection{Why Conventional Winning Tickets Are Hard to Find on ViTs? -- An Empirical Study} \label{weight-level-experiments-details}
We try to identify the winning tickets in ViTs with the same settings of conventional LTH~\cite{frankle2018lottery}. We implement the DeiT-Small~\cite{pmlr-v139-touvron21a-deit} in the experiments and obtain the sparse mask by pruning the pretrained model weights in layers of multi-head self-attention (MSA) and MLP blocks.
Following the conventional LTH, we train the models from scratch with the proper or random weight initialization and apply the same sparse mask.
The results of experiments implemented by DeiT-Small are shown in Table~\ref{weight-lth-1}. 

Although the model with proper initialization and low weight sparsity can match the accuracy of pretrained model, there is no accuracy difference (for the same weight sparsity) between the models trained from proper initialization (LTH) and random reinitialization (RR).
According to Table~\ref{weight-lth-1}, the performance of the LTH is lower than the performance of the RR with several different sparsity ratios, which means that the initializations are not as effective as in conventional LTH.
Besides, we import the rewinding technique to improve the accuracy of the models trained from proper initialization (LTH). We achieve more than 2.4\% in accuracy compared to the model trained from random reinitialization (RR) with 71\% weight sparsity. However, there is still a 5\% accuracy drop between the improved accuracy and the pretrain accuracy, which means the LTH is hard to match the pretrain accuracy even if the rewinding technique is used.
Thus, the subnetworks on weights that can be identified as winning tickets according to the conventional LTH are hard to find in ViTs.

Moreover, we also conduct the same experiments on a CNN where the original LTH is evaluated. We use a ResNet-20~\cite{he2016deep} on CIFAR-10~\cite{krizhevsky2009learning}, and the results are showed in the Table~\ref{weight-lth-2}.

The results on ResNet-20 show that there is a small accuracy drop between the model trained from proper initialization (LTH) and pretrained model even at high weight sparsity (e.g., 90.1\% vs. 90.3\% Top-1 accuracy at 83.2\% sparsity). 
The accuracy drop between the model trained from proper and random initialization is large (2.2\% $\sim$ 6.3\% Top-1 accuracy difference under the same sparsity), which shows the identification of the conventional winning tickets. Furthermore, the accuracy is improved obviously when the rewinding technique is used in the experiments (e.g., 88.9\% vs. 86.4\% Top-1 accuracy at 95.6\% sparsity ).

These experimental results guide us to revisit the structure of ViTs, which might explain the difficulty of the identification of conventional winning tickets on ViTs.
Unlike the CNNs convolutional kernel, which is an input-independent parameter of the static value.
The matrix multiplication in the MSA module brings interaction between tokens, which means that ViT closely depends on the representation of the input.
Due to the input-dependency characteristics of ViTs, the subnetwork produced by one sparse mask is difficult to generalize to the different input data.
Therefore, we decide to identify winning tickets at the input level instead of model weights for ViTs.

%%%%%%%%%%%%%%%%%%%%%%%%%
\begin{table}[t]
\centering
\resizebox{.999\columnwidth}{!}{

\begin{tabular}{ c | c | c  c c}
        \toprule
        \makecell{DeiT-Small \\ Weights Pruned}  & \makecell{Weight \\ Sparsity} & \makecell{LTH \\ Acc. (\%)} & \makecell{RR \\ Acc. (\%)} & \makecell{Acc. \\ Diff. (\%)} \\ \midrule
        None & 0\% & 79.9 & 79.9 & 0 \\ \midrule
        MSA  & 12\% & 79.6  & 79.5 & 0.1 \\ 
        MSA \& MLP  & 44\% & 77.2 & 77.3 & -0.1 \\ 
        {MSA \& MLP}$^{*}$ & 44\% & 78.1  & 77.6 & 0.5 \\ 
        MSA \& MLP  & 53\% & 76.5 & 76.6 & -0.1 \\
        MSA \& MLP$^{*}$  & 53\% & 78.6 & 76.8 & -0.2 \\
        % MSA \& MLP  & 71\% & 67.8 \textcolor{blue}{(-12.1)} & 67.7 \textcolor{blue}{(-12.2)} & 67.6 \textcolor{blue}{(-12.3)}  \\
        MSA \& MLP  & 71\% & 72.6 & 72.7 & -0.1 \\
        {MSA \& MLP}$^{*}$ & 71\% & 74.9 & 72.5 & 2.4 \\
        \bottomrule
        \end{tabular}
        }
\caption{
Results of experiments following conventional LTH settings. The results in the table are based on DeiT-Small~\protect\cite{pmlr-v139-touvron21a-deit} with ImageNet~\protect\cite{deng2009imagenet} dataset. The MSA indicates the multi-head self-attention layers, and the MLP indicates the MLP blocks in the transformer encoders. LTH Acc. denotes the accuracy of the winning tickets (i.e., trained with a sparse mask from the appropriate initialization); RR Acc. denotes the accuracy of the model trained by the same mask but from another random reinitialization; Acc. Diff. denotes the accuracy difference between LTH Acc. and RR Acc.; and the * in the table denotes that the rewinding technique is used in experiments.}
\label{weight-lth-1}
\end{table}

\begin{table}[t]    % these are lr=0.01, imp results
\centering
\resizebox{.999\columnwidth}{!}{

\begin{tabular}{ c |  c | c  c c}    % the rewind of RR is evaluated
        \toprule
         \makecell{ResNet-20 \\ Weights Pruned}  & \makecell{Weight \\ Sparsity} & \makecell{LTH \\ Acc. (\%)} & \makecell{RR \\ Acc. (\%)} & \makecell{Acc. \\ Diff. (\%)} \\ \midrule
        None & 0\% & 90.3 & 90.3 & 0  \\ \midrule
        All Conv layers & 83.2\% & 90.1 & 87.9 & 2.2\\
        All Conv layers & 91.4\% & 89.4 & 85.7 & 3.7\\ 
        {All Conv layers}$^{*}$ & 91.4\% & 89.8 & 85.8 & 4.0 \\ 
        All Conv layers & 94.5\% & 87.5  & 83.9 & 3.6 \\ 
        {All Conv layers}$^{*}$ & 94.5\% & 89.4 & 83.9 & 5.5 \\ 
        All Conv layers & 95.6\% & 86.4  & 83.4 & 3.0 \\ 
        {All Conv layers}$^{*}$ & 95.6\% & 88.9  & 83.3 & 6.3 \\
        % All Convolution Layers & 59.0\% & 90.1 \textcolor{blue}{(-0.2)} & 89.3 \textcolor{blue}{(-1.0)} & 89.8 \textcolor{blue}{(-0.5)}  \\
        \bottomrule
        \end{tabular}
        }
\caption{
 The results in the table are based on all convolutional layers in ResNet-20~\protect\cite{he2016deep} on CIFAR-10~\protect\cite{krizhevsky2009learning}.}
\label{weight-lth-2}
\end{table}
%----------------------

%-------------------------------------------------------------------------------------------------------------------------------

\subsection{Patch Redundancy}
The Transformer is first introduced to the computer vision research area by~\cite{carion2020end}. In this work, each image is first split into a sequence of fix-sized and non-overlapping patches, which are then projected to a sequence of tokens with embeddings before being fed into the transformer blocks.
Each token is dependent on a unique patch of the input image according to the positional embedding~\cite{pmlr-v139-touvron21a-deit}. Furthermore, the tokens are independent of each other in the self-attention module of transformer blocks, which means the local information of each patch is not utilized.
%***Why to do token pruning*******
%****1. has redundancy*******
The reason why there exists redundancy in patches is that the input images are evenly segmented. Therefore, many patches may contain only background information and lack object information. Those patches contribute less to the classification task, so they can be removed without affecting the accuracy.
%****2. can reduce more flops*******
%We reduce token redundancy over other dimensions (e.g. weight pruning and token channel pruning) because it is the most efficient way to reduce the computation.
%---------------------------------------------------------------------------------------
Our choice of patch removal over weight pruning can be illustrated from two perspectives: 1) From the computation standpoint, it is the most efficient way to reduce model complexity. As shown in \cite{kong2022spvit}, under the same sparsity ratio, patch removal can reduce overall computation more than pruning channels and attention heads.
%****3. data dependant *******
2) From a vision interpretability standpoint, weight pruning changes the model itself, which is input-independent. However, each image contains information in a different location and size. It is not possible to obtain one sparse model that is valid for all images. Patch removal can obtain different input sparsity for each image, which can significantly reduce the training and testing computation without reducing the model's capability.
%---------------------------------------------------------------------------------------

%***How to do token pruning*******
Previous works~\cite{liang2022evit,rao2021dynamicvit} use some additional optimization methods to identify and remove inattentive tokens at specific stages of the model. They argue that the tokens cannot be pruned at an early stage because: 1) The attention maps are unreliable for token removal. 2) Tokens may contain more global information, making it difficult to identify uninformative tokens. 
These difficulties have discouraged them from applying their methods at early stages, which creates a barrier to optimization. Therefore, it is important for us to identify the important tokens as early as possible in ViTs. Also, it is intuitive for us to maximize the reduction in computation by removing the inattentive image patches before feeding them into the first transformer block.

To achieve such a purpose, we insert a ticket selector to evaluate image patches and remove less important ones.
Since the self-attention module in ViTs focuses on local information in early layers and global information in deeper layers, the ticket selector evaluates the importance of image patches according to the attention maps extracted from multiple depths of layers.
Therefore, the selected image patches are the most attentive and informative ones, which can be defined as the winning tickets. A more rigorous definition of winning tickets is provided in the definition section.

\subsection{Notations}
In this section, we create notations and functions listed in Table~\ref{tab:notation} with detailed descriptions to illustrate our theory.
%%%%%%%%%%%%%%%%%%%%%%%%%%%%%%%%%%%%%%%%%%%%%%%%%%%%%%%%%%%%%%%%%%
% take-care: detailed to talk about the 'p' in ticket selector
%%%%%%%%%%%%%%%%%%%%%%%%%%%%%%%%%%%%%%%%%%%%%%%%%%%%%%%%%%%%%%
\begin{table}[t]
\centering
\resizebox{.99\columnwidth}{!}{
\begin{tabular}{l | l}
    \toprule
    \textbf{Notation}   &   \textbf{Description} \\
    \midrule
    
    % $T$ &  $T$ is the total number of training epochs.  \\ \midrule 
    $T$ &  The total number of training epochs.  \\ \midrule 
    
    $F$ &  The total number of fine-tuning epochs. \\ \midrule
    
    $f_{t}(\cdot)$ &  Model trained from scratch for $t$ epochs. \\ \midrule 
    
    $f^{\Omega}(\cdot)$ &  Model trained with the full image. \\ \midrule 
    
    $f^{L}(\cdot)$ &  Model trained with the winning tickets. \\ \midrule 
    
    $f^{R}(\cdot)$ &  Model trained with randomly selected patches. \\ \midrule 
    
    % $p$ & $p$ is the token identification module in ticket selector, which is mainly based on the attention map and used to remove less important tokens. \\ \midrule
    $p$ & Token identification module\\ \midrule
    
    $m$ & The sparse mask. \\ \midrule
    % $m \in {\{0,1\}}^{|x|}$ is generated by fine-tuned model containing $p$. \\ \midrule
    
    $m^{\prime}$ & The random sparse mask. \\ \midrule
    % $m^{\prime} \in {\{0,1\}}^{|x|}$ is generated randomly. \\ \midrule
    
    % $s$ &  $s$ is the sparsity ratio, which is defined as the percentage of removed image patches in the ViT model. \\ 
    $s$ &  Sparsity ratio.\\ 
    
    \bottomrule

\end{tabular}
}
\caption{Summary of notations and functions.}
\label{tab:notation}
\end{table}

%%%%%%%%%%%%%%%%%%%%%%%%%%%%%%%%%%%%%%%%%%%%%%%%%%%%%%%%%%%%%%%%%%%%%%%%%%%%%%%%%%%%%%%%%%%%%%%%%%%%%%%%%%%%%%%%%%%%%%%%%%%%%%%%

Consider a Transformer based network function $f(\cdot)$ that is denoted as $f(x)$ when fed with the image $x$. We define the following settings:

%%%%%%%%%%%%%%%%%%%%%%%%%%%%%%%%%%%%%%%%%%%%%%%%%%%%%%%%%%%%%%%%%%%%%%%%%%%%%%%%%%%%%%%%%%%%%%%%%%%%%%%%%%%%%%%%%%%%%%%%%%%%%%%%
\begin{itemize}[leftmargin=*]

\item \emph{Pretraining:} We train the network $f^{\Omega}_{0}(\cdot)$ with the full image $x$ for $T$ epochs, arriving at $f^{\Omega}_{T}(\cdot)$.

\item \emph{Lottery Ticket:} We directly apply mask $m$ to input image patches, resulting in $x\odot m$ as winning tickets and the network function $f^{L}(\cdot)$.

\item \emph{Random Choice:} We directly apply mask $m^{\prime}$ to input image patches, resulting in $x\odot m^{\prime}$ as the subset of randomly selected image patches and the network function $f^{R}(\cdot)$.

\end{itemize}
%%%%%%%%%%%%%%%%%%%%%%%%%%%%%%%%%%%%%%%%%%%%%%%%%%%%%%%%%%%%%%%%%%%%%%%%%%%%%%%%%%%%%%%%%%%%%%%%%%%%%%%%%%%%%%%%%%%%%%%%%%%%%%%%

\subsection{Definition of the \xs{Data Level} Lottery Ticket Hypothesis for ViTs}\label{definition}

According to our previous discussion, we have identified the feasibility of implementing the sparse ViTs by removing image patches at the input level instead of removing tokens at different stages inside the ViT architectures. In the following subsections, we propose the definition of the \xs{Data Level} Lottery Ticket Hypothesis for Vision Transformers, and the principles of identifying the winning tickets for the ViTs. Additionally, for the mainstream ViT variants, we demonstrate the methodology for obtaining the corresponding winning tickets, which shows the universality and validity of our proposed definition.

\textbf{The \xs{Data Level} Lottery Ticket Hypothesis for Vision Transformers.} \emph{For a ViT model and a non-trivial sparsity ratio, there exists a \xs{data level} subset that contains the most attentive image patches that -- when trained in isolation -- can match\footnote{Accuracy drop between model trained with winning tickets and pretrain model is approximately within 0.5\%.~\cite{zhu2017prune,frankle2018lottery,zhou2019deconstructing}} the accuracy of the model that is trained with the full image while showing a clear advantage in accuracy compared to the model that trained with a randomly selected subset with the same sparsity ratio for the same or less number of epochs.}

In our definition of the \xs{Data-}LTH-ViTs, the subset of the input image patches is unique, in which each of the input images has a corresponding fixed mask that maintains a fixed topology of the dataset regardless of the sequence of the input images to the network. In order to achieve this, we make a reasonable analogy to the conventional LTH that uses the prior knowledge (i.e., the pretrained weights) to generate this unique mask.

Based on the pretrained weights, we import the additional module $p$ that can identify the importance of each token according to the attention map and build the ticket selector as $f^{\Omega}_{T}(\cdot; p)$. More specifically, \ding{172} the token identification module $p$ is inserted before $4^{th}$, $7^{th}$, $10^{th}$ layers. \ding{173} The module $p$ sorts the tokens according to the [CLS] token~\cite{dosovitskiy2020vit}, which is allowed to interact with all the other tokens in the attention map, and only keeps the top-k most informative tokens. \ding{174} The attentive image patches can be indexed according to positional embeddings of the remaining tokens generated by the last module $p$.
After the attentive image patches are selected, we further explore the identification of the winning tickets in these selected image patches. We train the model $f^{L}_{0}(\cdot)$ from scratch only with the subset of selected image patches $x\odot m$ for $T$ epochs and evaluate the performance of the model $f^{L}_{T}(\cdot)$ being fed with the same subset $x\odot m$ as $f^{L}_{T}(x\odot m)$.

In the conventional LTH, the irreplaceable weight initialization value of the winning tickets makes up the unique winning property for the LTH. The performance of the subnetworks drops drastically when the initialization changes. In our proposed \xs{Data-}LTH-ViTs, the identified winning tickets have similar but not quite the same characteristic. The similarity is reflected in the fact that the winning tickets must use the same subset, and the difference is that this subset is within the input data. When the subset $x\odot m$ is identified, the performance should surpass its randomly selected counterpart $x\odot m^{\prime}$ before calling it the winning ticket.
Thus, in our definition, it is necessary for us to explore the performance of the model trained with randomly selected image patches to validate the effectiveness and uniqueness of the winning tickets.

\subsection{The Identification of the Winning Tickets}

To make a valid identification of the winning ticket, three key evaluations should be performed: \ding{172} the test accuracy of the ViT $f^{L}_{T}(x\odot m)$, which is using the subset of attentive image patches $x\odot m$ for each \emph{test} image $x$, \ding{173} the test accuracy of ViT $f^{R}_{T}(x\odot m)$, which is using identical test image subset $x\odot m$ for testing, \ding{174} the test accuracy of the ViT $f^{\Omega}_{T}(x)$ using full input image patches. By cross-comparing the results of \ding{172} \ding{173} and \ding{174}, we can validate if the tickets are unique to be confidently identified as the winning ticket. When the test accuracy of $f^{L}_{T}(x\odot m)$ can match the pretrained ViT $f^{\Omega}_{T}(x)$, and there exists a clear accuracy gap between $f^{L}_{T}(x\odot m)$ and $f^{R}_{T}(x\odot m)$, the subset of the input patches $x \odot m$ can be considered as the winning ticket.

% \textbf{Analogical similarity between conventional LTH and the proposed \xs{Data-}LTH-ViTs.}
% Both conventional LTH and \xs{Data-}LTH-ViTs 
% \ding{172} attempt to identify the subsets of weights/inputs that can achieve similar accuracy to the model trained with full weights/inputs sets;
% \ding{173} need to use prior knowledge (i.e. a pretrained model) to identify the winning tickets;
% \ding{174} identify the winning tickets with the requirement of matching the accuracy of pretrained model;
% \ding{175} import randomness into the subsets to validate the uniqueness and effectiveness of the winning tickets.

\subsection{Methodology}

Different variants of ViT are imported to demonstrate the effectiveness and generalization ability of our definition. The main architecture we use in this paper is DeiT~\cite{pmlr-v139-touvron21a-deit}. To illustrate the integrity of our definition, we also explore the lottery ticket hypothesis for other ViT variants: Swin~\cite{liu2021Swin} and LV-ViT~\cite{jiang2021all-lvvit}, with some adjustments during the training and testing progress to match their model architecture.
We keep using DeiT-Small as a ticket selector in experiments on those variants of ViTs.

\textbf{Vanilla ViT.} For the \emph{Lottery Ticket}, we use the model DeiT-Small and DeiT-Tiny~\cite{pmlr-v139-touvron21a-deit} to validate the existence of winning tickets. We test our proposed \xs{Data-}LTH-ViTs definition using DeiT-Tiny and DeiT-Small.
The overview of our method is shown in Figure~\ref{figure-overview-of-training-testing-pipeline}.
\ding{172} We pretrain the model to get $f^{\Omega}_{T}(x)$, which is the same as conventional LTH. \ding{173} We incorporate the token identification module $p$ into the model to build the ticket selector $f^{\Omega}_{T}(x; p)$. \ding{174} We can use the ticket selector to produce the index of inattentive tokens for the mask $m$ to build the winning tickets as $x\odot m$. \ding{175} We train the model only with the winning tickets for the same iterations as pretrained model.
We evaluate the performance of the model trained with the full image $x$ or winning tickets $x\odot m$ to see if the test accuracy of the lottery ticket can match with the model trained and tested using full image patches. For the \emph{Random Choice}, we replace the winning tickets with randomly selected patches with the same sparsity ratio during training, and the accuracy of the \emph{Random Choice} is evaluated by using a full image or identical test image patches that are generated by $f^{\Omega}_{T}(x; p)$ to ensure fairness.

\begin{figure}[t]
    \centering
    \includegraphics[width=0.47\textwidth]{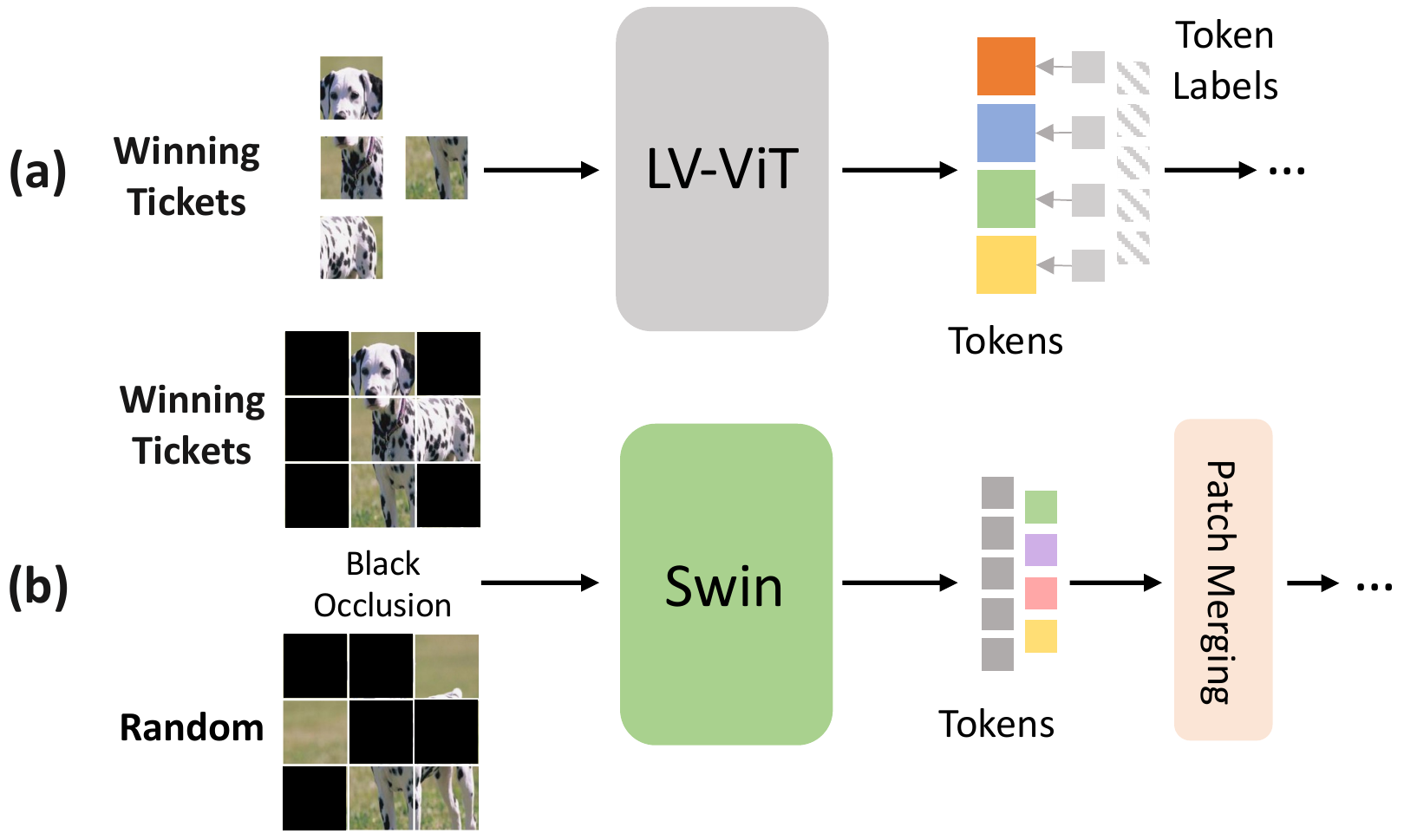}
    \caption{Adjustments on two variants of ViTs. (a) the LV-ViT, remove corresponding token labels according to the patch removal during the training process; (b) the Swin Transformer, use all-zero values to occlude the inattentive patches to maintain the input size and window size during the training process.}
    \label{figure-2-variants-vits}
\end{figure}

\textbf{LV-ViT.} We use the LV-ViT-Small~\cite{jiang2021all-lvvit} as the main architecture in experiments. 
According to the design of LV-ViT~\cite{jiang2021all-lvvit}, the token labeling technique is introduced to improve the model performance of the model. This method takes advantage of all tokens to compute the training loss in a dense manner instead of computing loss on the additional learnable class token~\cite{jiang2021all-lvvit}. To improve the recognition on the token level, each token in the LV-ViT is assigned with an additional label as location-specific supervision. Therefore, as Figure~\ref{figure-2-variants-vits} (a) shows, we apply the same sparse mask on those labels as it is on image patches to correct the computation of training loss because the number of tokens is reduced when input image patches are applied with the sparse mask.

\textbf{Swin Transformer.} We mainly implement the Swin-Tiny~\cite{liu2021Swin} in our experiments. 
We turn the pixel values of those inattentive patches to zero as the substitute for removal, and the pixels of the winning tickets produced by the ticket selector remain constant, as shown in Figure~\ref{figure-2-variants-vits} (b).
The purpose of using black occlusion rather than directly removing them is due to the introduction of shifted windows in the self-attention module of Swin Transformer~\cite{liu2021Swin}.
If we completely remove the inattentive patches, we need to revise the window size according to the number of removed image patches when we reshape the remaining image patches into another small image in the square because the window size should be divisible by the input image size. Additionally, smaller input resolution and window size, along with the change of relative patch positions, would hurt the performance of the Swin.
Therefore, to maintain the performance, the inattentive image patches are obscured by black color rather than directly removing them.

\section{Experiments and Results} \label{experiments}

\subsection{Experimental Setups}
In this paper, all of the models are trained on the ImageNet~\cite{deng2009imagenet} with approximately 1.2 million images in the training set and all of the results of accuracy are tested on the 50k images in the testing set. The image resolution in training and testing is 224 × 224, and the experiments are conducted with different sparsity ratios.
As for the models of DeiT-Small and DeiT-Tiny~\cite{pmlr-v139-touvron21a-deit}, we use the same architecture for both the ticket selector and training model in the experiments. And as for the LV-ViT~\cite{jiang2021all-lvvit} and Swin~\cite{liu2021Swin}, we introduce the DeiT-Small~\cite{pmlr-v139-touvron21a-deit} as the ticket selector to verify the generalization of the winning tickets. For different variants of models, we conduct a series of experiments with several sparsity ratios. For the training strategies and optimization methods, we follow the principles proposed by the original papers of DeiT~\cite{pmlr-v139-touvron21a-deit}, Swin~\cite{liu2021Swin} and LV-ViT~\cite{jiang2021all-lvvit}. For the testing process, all the models are tested in two different ways. We test the model with the winning tickets produced by the same ticket selector as it is in training progress or with the full test images, which can demonstrate the identification of the winning tickets. All the experiments are conducted on the NVIDIA A100 with 8 GPUs.

\subsection{Main Results}
\begin{figure}[t]
\centering
    \includegraphics[width=0.45\textwidth]{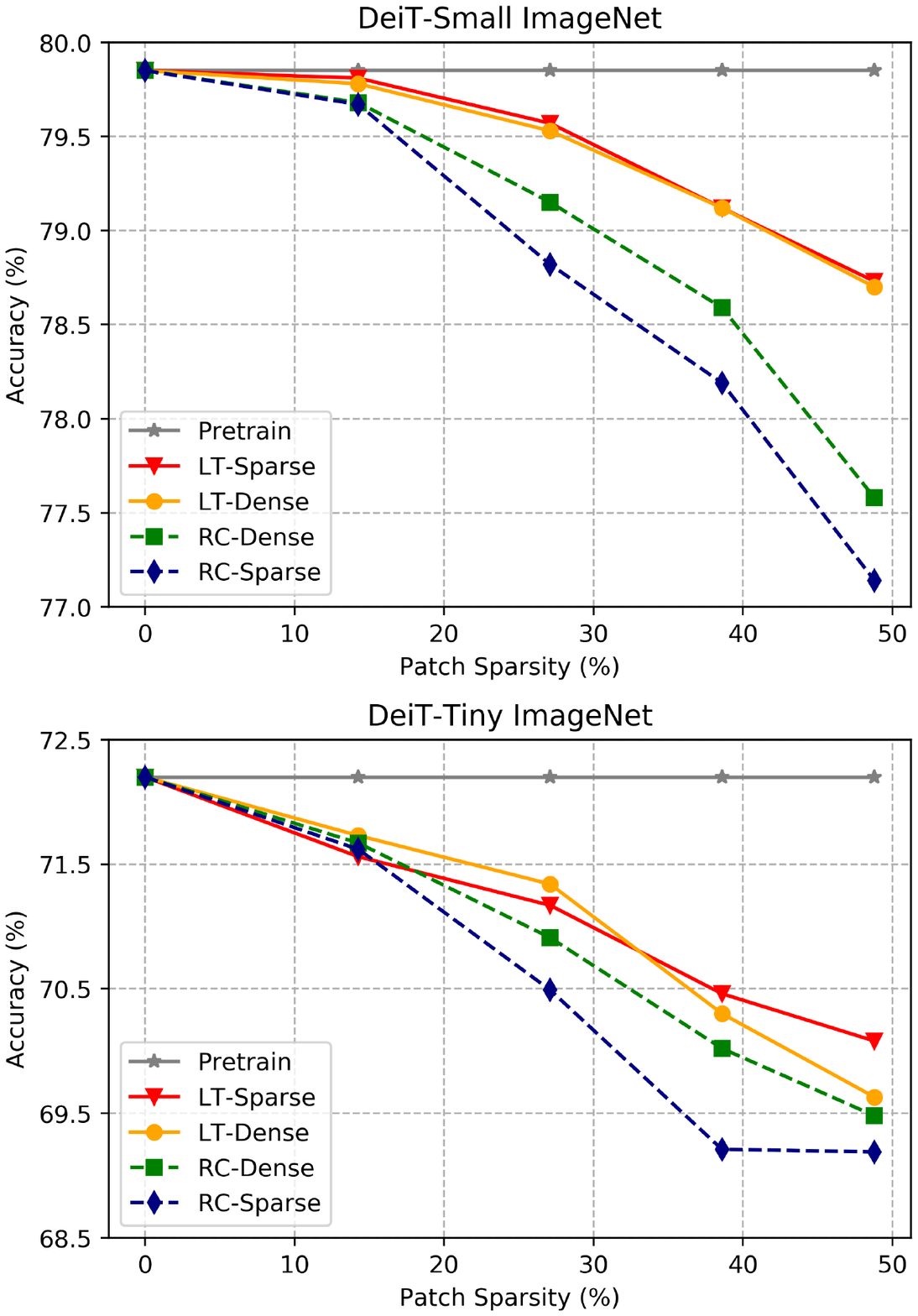}
\caption{Top-1 accuracy vs. Patch sparsity on ImageNet. We show the lottery ticket experiments with DeiT Small and DeiT Tiny on ImageNet dataset and compare the experimental results with the pretrain accuracy. The LT and RC denote the model trained with \emph{Lottery Ticket} and \emph{Random Choice}, and the Dense and Sparse indicates the model is tested with full image and winning tickets. }
\label{figure-deit-small-and-tiny-results}
\end{figure}

We report the main results using the models of DeiT-Small and DeiT-Tiny~\cite{pmlr-v139-touvron21a-deit} in Figure~\ref{figure-deit-small-and-tiny-results} with the top-1 accuracy vs. patch sparsity ratio curves plotted. The figure shows the results of two models with dense or sparse inference in four different colored lines that vary with four different patch sparsity ratios. The Dense and Sparse inference means that we evaluate the model with the full test image patches and the winning tickets, respectively. And the LT (\emph{Lottery Ticket}) and RC (\emph{Random Choice}) denote that the model is trained with winning tickets and the subset of randomly selected patches, respectively. It is obvious in Figure~\ref{figure-deit-small-and-tiny-results} that the LT model colored with red and orange can achieve better accuracy than the RC model colored in green and blue. The accuracy drops between four lines become large with the increase in the sparsity ratio.
Furthermore, according to the results of DeiT-Small, the LT model can match the accuracy of the pretrained model colored in grey with Sparse inference because the accuracy drops between them are less than 0.5\% when the sparsity is 27\% and less than 1\% even the sparsity is about 40\%. As for the model of DeiT-Tiny, the accuracy drops between the LT model and pretrained model at the two smallest patch sparsity ratios are less than 1\%, and the difference between the LT model and RC model is noticeable as patch sparsity increases. Besides, we can find that, in the figure, models trained and tested with winning tickets have the biggest advantage over others at the largest patch sparsity ratio. Thus, based on these observations, the identification of the winning tickets in normal ViTs is demonstrated.

\subsection{Generalization Analysis}
% CARE-todo: special, with proposed techniques, hard to find on CNNs (results in appendix)
Besides the DeiT model, we import LV-ViT~\cite{jiang2021all-lvvit} and Swin Transformer~\cite{liu2021Swin} to demonstrate the generalization of the winning tickets as shown in Table~\ref{table-other-kinds-of-ViT}. The DeiT-Medium has a similar structure to DeiT-Small, while the dimension of embeddings in this model is changed to 576, and the number of heads is changed to 9. As for the Swin Transformer, we use the Swin-Tiny model with two different patch sparsity ratios to explore the identification of the winning tickets on the Swin Transformer and use all-zero values to occlude the inattentive patches to maintain the input size and window size during the training process. For LV-ViT, we train the LV-ViT-Small with token labels proposed by LV-ViT during the training process. All the accuracy of LT models can match the accuracy of the pretrained model at the given patch sparsity ratios and there are noticeable accuracy drops between LT models and RC models, which demonstrate the identification of the winning tickets. Thus, the generalization of the winning tickets through our definition of LTH for ViTs can be verified.

We also try to identify the winning tickets on CNNs with our proposed method, we use the ResNet-50~\cite{he2016deep} to explore the identification of the winning tickets, and we use all-zero values to occlude the inattentive patches rather than cropping the image in the training progress. The experimental results show that CNNs is difficult to identify the new kind of winning tickets because the LT models cannot match the pretrain accuracy, and there is no accuracy drop between the LT models and RC models.

\begin{table}[t]
\centering
\resizebox{.99\columnwidth}{!}{
\begin{tabular}{ c| c c c c |c }
\toprule
Models & Swin-T & Swin-T & LV-ViT-S & DeiT-M & ResNet-50 \\ \midrule
Pretrain Acc. (\%) &  81.2 & 81.2  & 83.3 & 83.3 & 77.34 \\ \midrule
Patch Sparsity (\%) &  27.1 & 48.8 & 38.6 & 14.3 & 27.1\\ \midrule
LT-Sparse Acc. (\%)   &  - & -   & 82.4 & 81.1 & - \\ 
LT-Dense Acc. (\%)   &  80.95 & 80.63   & 82.5 & 81.2 & 76.47 \\
RC-Dense Acc. (\%)   &  79.85 & 78.37   & 82.1 & 80.6 & 76.86 \\
RC-Sparse Acc. (\%)   &  - & -   & 81.9 & 80.7 & - \\ 
\bottomrule
\end{tabular}}
\caption{
Identification of winning tickets on DeiT-Medium, Swin-Tiny and LV-ViT-Small.
The DeiT-Medium is similar to DeiT-Small but with 576 embedding dimensions and 9 heads.
The input size is maintained as Dense in Swin-Tiny and ResNet-50 because of the zero-value occlusion.
}
\label{table-other-kinds-of-ViT}
\end{table}

\section{Conclusion, Limitation, and Societal Impact}

In this paper, we present the first Data Level Lottery Ticket Hypothesis for Vision Transformers (Data-LTH-ViTs), which inherits the essential concepts and exhibits generalization to the conventional LTH that is mainly discussed for the weights in CNNs. We conduct empirical studies to explain why conventional lottery tickets are hard to find on ViTs, and propose our novel definition of Data-LTH-ViTs and the principles for the identification of the winning tickets on the input data of ViTs. 
We demonstrate that there exists a subset of input image patches (i.e., winning tickets of Data-LTH-ViTs) that are capable of training a ViT and its multiple variants to achieve similar accuracy as the ones trained using the full dataset.
% This paper mainly focuses on image classification tasks while there are ViTs variants designed for other tasks such as segmentation and detection, etc. We leave these tasks for our future work.
% The research in this paper is scientific in nature and we do not envision it generating any negative societal impact.

\bibliographystyle{named}
\bibliography{ijcai23}

% \clearpage
% \newpage
% \appendix
% \input{sections/6_appendix}

% \clearpage
% \newpage
% \clearpage
% \input{sections/7_resubmission}

\end{document}